\documentclass[runningheads]{llncs}

\usepackage[T1]{fontenc}
\usepackage{graphicx}
\usepackage{amsmath,amssymb,amsfonts}
\usepackage{algorithmic}
\usepackage{textcomp}
\usepackage{xcolor}
\usepackage{subfig}
\usepackage{gensymb}
\usepackage[numbers]{natbib}
\usepackage{caption}
\usepackage{svg}
\usepackage{mwe}
\usepackage{amsfonts}
\usepackage{amssymb}
\usepackage{verbatim}
\usepackage{amsbsy}
\usepackage{siunitx}
\usepackage{tabularx, booktabs}
\usepackage{soul}
\usepackage{tikz}
\usetikzlibrary{calc}
\usepackage{float}

\begin{document}

\title{UruBots Autonomous Cars Team One Description Paper for FIRA 2024}

\author{Pablo Moraes\inst{1} \and Christopher Peters\inst{2} \and Any Da Rosa\inst{1} \and Vinicio Melgar\inst{1} \and Franco Nuñez\inst{1} \and Maximo Retamar\inst{1} \and William Moraes\inst{1} \and Victoria Saravia\inst{1} \and Hiago Sodre\inst{1} \and Sebastian Barcelona\inst{1} \and Anthony Scirgalea\inst{1} \and Juan Deniz\inst{1} \and Bruna Guterres\inst{1} \and André Kelbouscas\inst{1} \and Ricardo Grando\inst{1} }

\institute{Technological University of Uruguay, UTEC, 
Ostfalia University of Applied Sciences }  

\authorrunning{UruBots AC One et al.}
\titlerunning{UruBots AC One}

\maketitle  
\begin{abstract}
This document presents the design of an autonomous car developed by the UruBots team for the 2024 FIRA Autonomous Cars Race Challenge. The project involves creating an RC-car sized electric vehicle capable of navigating race tracks with in an autonomous manner. It integrates mechanical and electronic systems alongside artificial intelligence based algorithms for the navigation and real-time decision-making. The core of our project include the utilization of an AI-based algorithm to learn information from a camera and act in the robot to perform  the navigation. We show that by creating a dataset with more than five thousand samples and a five-layered CNN we managed to achieve promissing performance we our proposed hardware setup. Overall, this paper aims to demonstrate the autonomous capabilities of our car, highlighting its readiness for the 2024 FIRA challenge, helping to contribute to the field of autonomous vehicle research.

\end{abstract}

\section{Introduction}

The development of autonomous vehicles marks an advancement in transportation technology and competitions such as the FIRA Autonomous Cars Race Challenge helps on the development and fostering of this area of study. Here we present our autonomous car, designed for the 2024 FIRA Autonomous Cars Race Challenge. This project focuses on creating a fully autonomous, RC-car sized electric vehicle capable of navigating race tracks environments with good precision and in a small amount of time.

Our vehicle integrates mechanical and electronic systems with learning algorithms. By using many sensors and using this information to train and feedback our model, our car manages to achieve environmental perception, being able to follow a proposed track and even avoid obstacles. The software architecture is designed for real-time processing with a dedicated board to run the AI model.

Overall, this document presents a detailed examination of both the software and hardware components of our autonomous car. The hardware section covers the mechanical design, electronic systems, and sensor integration, while the software section details the algorithms, data processing techniques, and how our vehicle perform the decision-making. We performed an evaluation on an example of of race track, where our vehicle was able to race a track of 13 meters in less than 20 seconds.

\section{Construction}

\subsection{Hardware}

To introduce the construction specifications of this vehicle, it is possible to view its actual physical dimensions in table 1 below. 

The vehicle's hardware was fully developed according to the competition's needs, with fully defined dimensions. The description of its parts are included in the following subsection with the description of the components and their functions for operation. An initial image of the vehicle in its configuration phase for operation can be seen in Figure 1.

\begin{table}[h!]
\centering
\begin{tabular}{|>{\bfseries}l|c|}
\hline
\textbf{Dimension} & \textbf{Size} \\
\hline
Length & 330 mm \\
\hline
Width & 190 mm \\
\hline
Height & 170 mm \\
\hline
Weight & 2.2 kg \\
\hline
\end{tabular}
\caption{Specifications of the Autonomous Car}
\label{tab:car_specifications}
\end{table}

\begin{figure}
    \centering
    \includegraphics[width=0.6\linewidth]{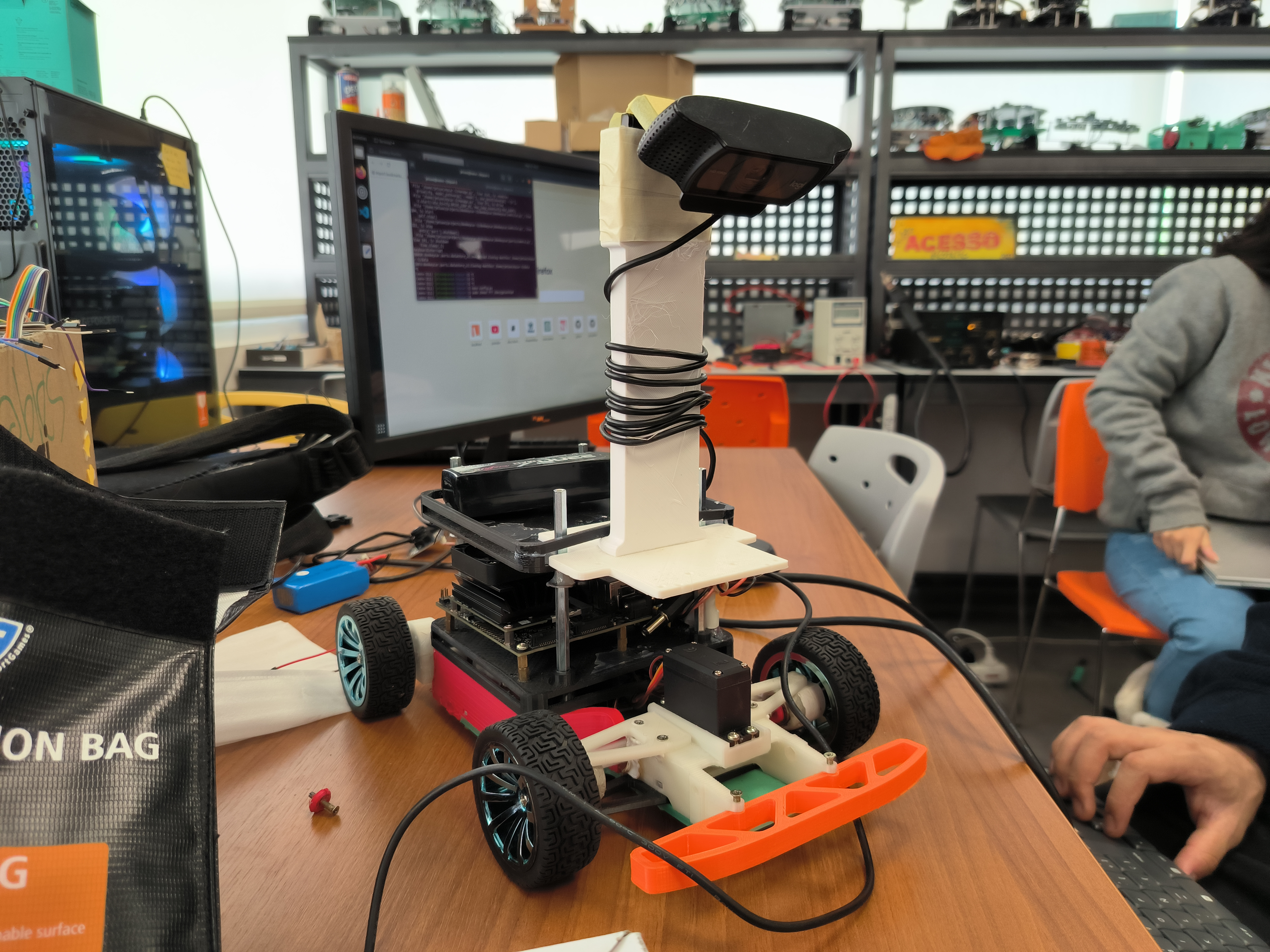}\captionsetup{justification=centering}\label{fig:car}]
    \caption{\parbox{0.8\linewidth}{The Autonomous car during the construction phase.}}
\end{figure}

Our autonomous vehicle is characterized by a design that combines PLA+ in its chassis to ensure both strength and lightness, having also a low cost bias. Its computational core lies in a Jetson Nano 4GB, backed by a 3D-printed casing for protection. Equipped with a Logitech C920 HD Pro camera for real-time vision, and powered by two 11.1V LiPo batteries, the vehicle ensures real time performance.

The vehicle's locomotion is achieved through 6V DC motors and an SG5010 servo motor, controlled respectively by a PCA9685 and an L298N H-bridge. Additionally, a TP-Link TX20 USB WiFi adapter is incorporated for wireless connectivity. The vehicle uses 19mm bearings and soft iron shafts for the Ackerman system. A XL4016 step-down converter is also included to power the Jetson Nano. These components are interconnected following a specific scheme, ensuring coordinated and efficient operation throughout all autonomous vehicle operations.

\begin{figure}[H]
  \centering
  \subfloat[Chassis: PLA+.\label{fig:chassis}]{
    \begin{minipage}[c][\width]{0.18\textwidth}
       \centering
       \includegraphics[width=\textwidth]{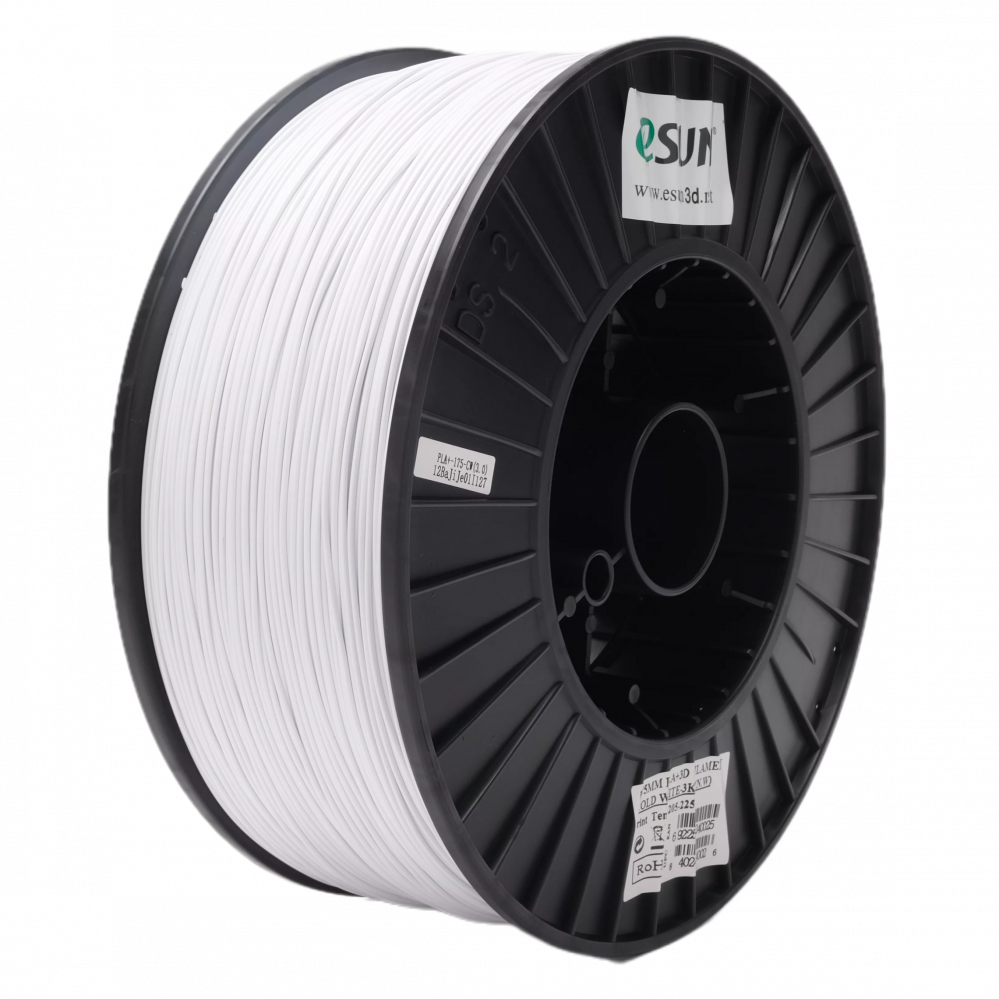}
    \end{minipage}}
  \hfill     
  \subfloat[Jetson Nano: 4GB.\label{fig:jetson}]{
    \begin{minipage}[c][\width]{0.18\textwidth}
       \centering
       \includegraphics[width=\textwidth]{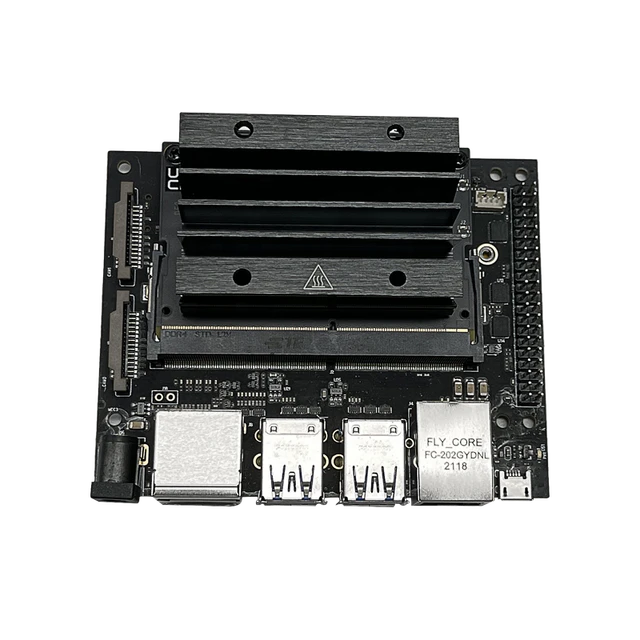}
    \end{minipage}}
  \hfill     
  \subfloat[Camera: Logitech C920 HD Pro.\label{fig:camera}]{
    \begin{minipage}[c][\width]{0.18\textwidth}
       \centering
       \includegraphics[width=\textwidth]{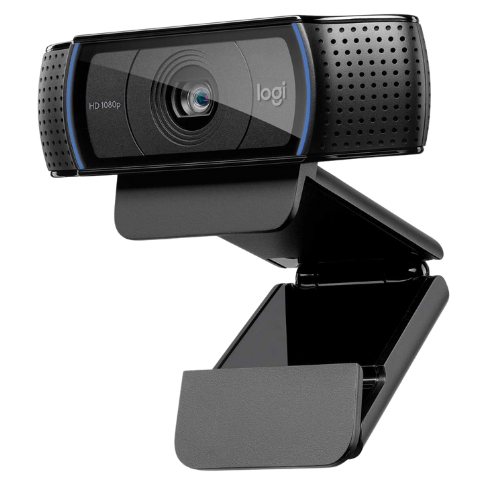}
    \end{minipage}}
  \hfill     
  \subfloat[Batteries: Two 11.1V LiPo.\label{fig:battery}]{
    \begin{minipage}[c][\width]{0.18\textwidth}
       \centering
       \includegraphics[width=\textwidth]{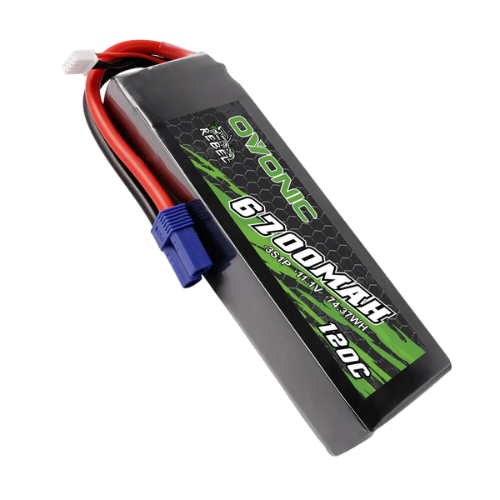}
    \end{minipage}}
  \hfill     
  \subfloat[DC Motors: 6V.\label{fig:motors}]{
    \begin{minipage}[c][\width]{0.20\textwidth}
       \centering
       \includegraphics[width=\textwidth]{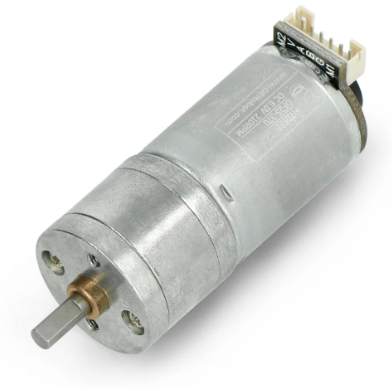}
    \end{minipage}}
  \hfill     
  \subfloat[Servomotor: SG5010.\label{fig:servo}]{
    \begin{minipage}[c][\width]{0.18\textwidth}
       \centering
       \includegraphics[width=\textwidth]{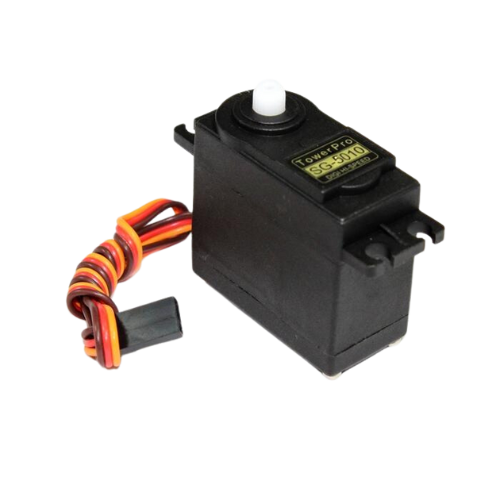}
    \end{minipage}}
  \hfill     
  \subfloat[Motor Control: PCA9685.\label{fig:control}]{
    \begin{minipage}[c][\width]{0.18\textwidth}
       \centering
       \includegraphics[width=\textwidth]{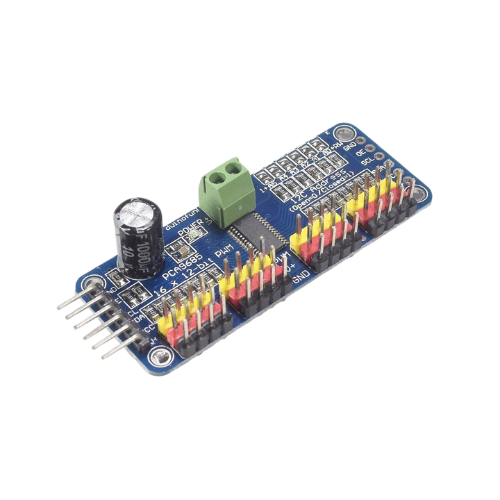}
    \end{minipage}}
  \hfill
  \subfloat[Motor Driver: L298N H-Bridge.\label{fig:bridge}]{
    \begin{minipage}[c][\width]{0.18\textwidth}
       \centering
       \includegraphics[width=\textwidth]{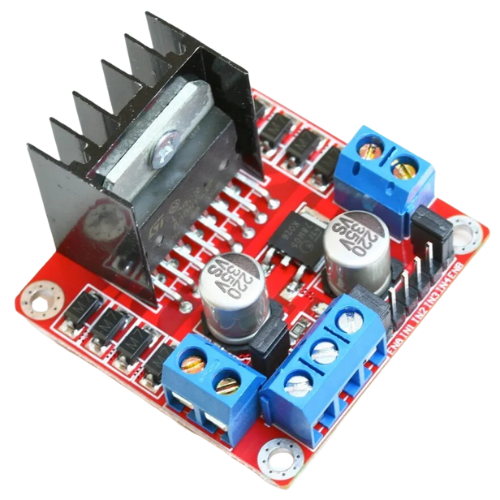}
    \end{minipage}}
  \hfill
  \subfloat[WiFi Adapter: TP-Link TX20 USB.\label{fig:wifi}]{
    \begin{minipage}[c][\width]{0.18\textwidth}
       \centering
       \includegraphics[width=\textwidth]{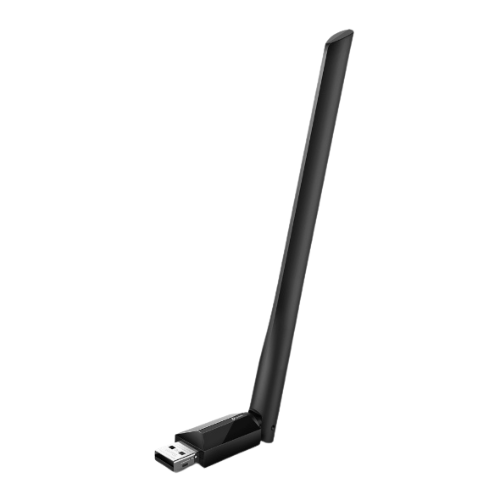}
    \end{minipage}}
  \hfill     
  \subfloat[Step Down XL4016: Power supply for Jetson Nano.\label{fig:stepdown}]{
    \begin{minipage}[c][\width]{0.18\textwidth}
       \centering
       \includegraphics[width=\textwidth]{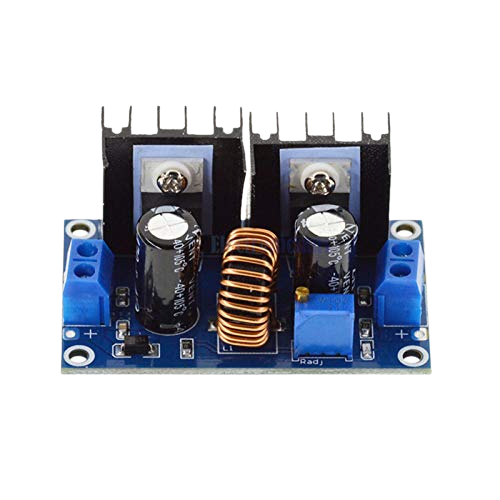}
    \end{minipage}}
  \hfill
  \subfloat[Bearings: 19mm.\label{fig:bearings}]{
    \begin{minipage}[c][\width]{0.18\textwidth}
       \centering
       \includegraphics[width=\textwidth]{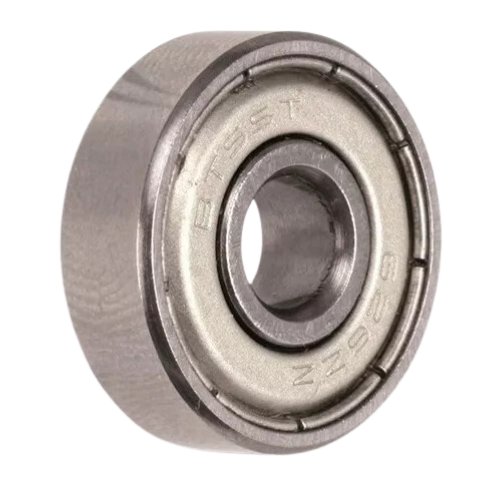}
    \end{minipage}}
  \hfill
  \subfloat[Ackerman System.\label{fig:ackerman}]{
    \begin{minipage}[c][\width]{0.18\textwidth}
       \centering
       \includegraphics[width=\textwidth]{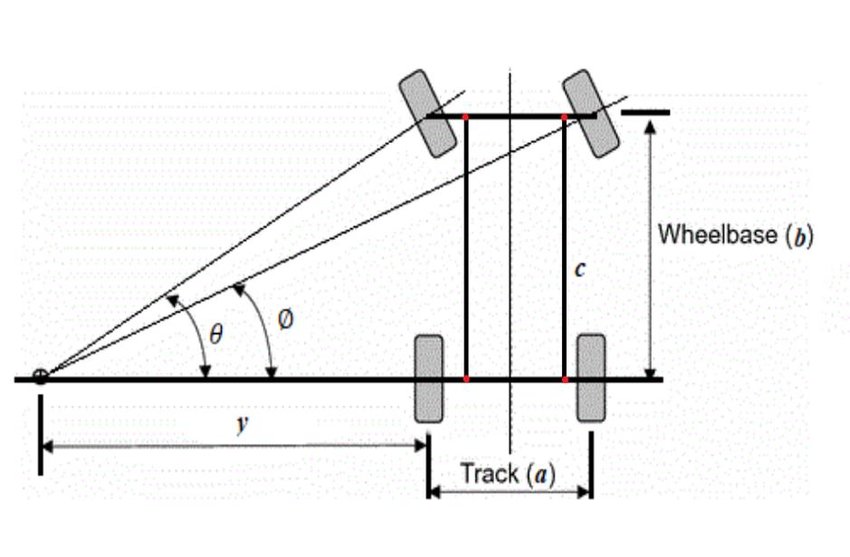}
    \end{minipage}}
  \hfill     
  \subfloat[Wheel shafts: Soft iron for the Ackerman system.\label{fig:shafts}]{
    \begin{minipage}[c][\width]{0.18\textwidth}
       \centering
       \includegraphics[width=\textwidth]{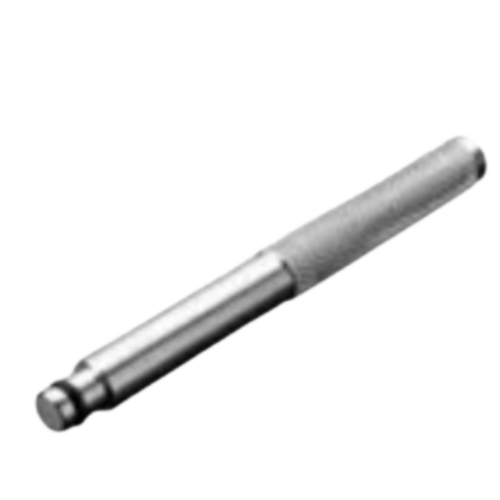}
    \end{minipage}}
  \caption{Components of our autonomous vehicle.}
\end{figure}

\begin{table}
    \centering
    \caption{Presented below is a tabulated enumeration of the components delineated within. }
    \label{tab:my_label}
    \begin{tabular}{|p{3cm}  |p{8cm}|} \hline 
           Component& Details\\ \hline 

           Jetson Nano 4gb& A compact, powerful single-board computer by NVIDIA with a quad-core ARM CPU, 128-core GPU, and 4GB RAM (Figure b).\\ \hline 
           Lipo Battery 11.1v 3s& These are 11.1V 3-cell lithium polymer batteries, commonly used in RC vehicles, drones, and other high-power, lightweight electronic devices (Figure d).\\ \hline 
           Logitech C920& The Logitech C920 is a high-definition webcam known for its sharp video and audio quality. (Figure c).\\ \hline 
           DC Motors 6v& These 6-volt DC motors are versatile and are used in robotics applications for their simplicity and torque-boosting system. (Figure e).\\ \hline 
           PCA 9685& The PCA9685 is a 16-channel, 12-bit PWM controller used for precise control of multiple servo motors and DC motors, connected to our H-bridge.(Figure g).\\ \hline 
           L298N Bridge& An H-Bridge is a circuit that allows control of the direction and speed of DC motor (Figure h) \\ \hline 
           XL4016 Buck Converter& The XL4016 is a DC-DC converter that reduces a high input voltage to a lower one. It is used to stably power our Jetson Nano. (Figure j).\\ \hline 
           SG5010 Servomotor& The SG5010 is a servo motor known for being affordable and useful in robotics. It has very precise positioning control. (Figure f).\\ \hline 
           TP-LINK TX20u USB Adapter& The TP-LINK TX20u USB Adapter enables our Jetson Nano, which lacks built-in Wi-Fi, to connect to wireless networks. (Figure i).\\ \hline
    \end{tabular}
\end{table}

\subsection{Software}

Our software system is based on the Donkey Car open-source framework. This framework was designed for building self-driving robotic vehicles using Python and the Raspberry Pi or for the Jetson Nano board. We used this framework to perform the steering and throttle of our vehicle on a Jetson Nano board. Our proposed workflow to achieve an autonomous car are described as follows:

Our vehicle was projected with a monocular camera pointed towards the front of the vehicle as shown in Figure 1. The onboard camera captures real-time images of the track, serving as the primary input for navigation. The Jetson Nano processes these images, utilizing CUDA support for intensive image processing. A CNN with a Keras Linear architecture, comprising five convolutional layers, analyzes the images and predicts the optimal action. 5000 images were recorded in our dataset from parts of the an example track that we used for our qualification environment. Predictions from the CNN are translated into commands for motor and steering control. These commands are sent via the Donkey Car control interface, integrated with the vehicle's hardware. PWM controllers manage motor speed and wheel direction.

A web interface allows real-time monitoring of the vehicle's status, including speed, direction, and camera feed. Flask, a Python micro web framework, powers this interface, running on the Jetson Nano. All navigation data (images, control commands, telemetry) are logged for analysis. Pandas and Matplotlib facilitate data analysis and visualization, essential for refining the CNN model. Image capture is configured for real-time navigation, while the Jetson Nano processes images using the TensorFlow-based CNN. The PWM controller translates CNN commands into motor and steering control. The web interface offers real-time control and monitoring from any network-connected device. This comprehensive software system enables our autonomous vehicle to navigate efficiently and precisely in controlled environments, meeting the requirements of the FIRA competition.


Our convolutional model summary can be seen in the Figure 2. It has a total amount of five convolutional layers followed by some dense layers that at the end provide the steering an throttle for the vehicle; all of them with added dropuot layers. The 5000 images from our collected dataset were resized to 120x160 pixels and used to train for a total amount of 100 epochs. The model trained then was deployed and used then to control the vehicle.

\begin{figure}[httb]
    \centering
           \includegraphics[scale=0.4]{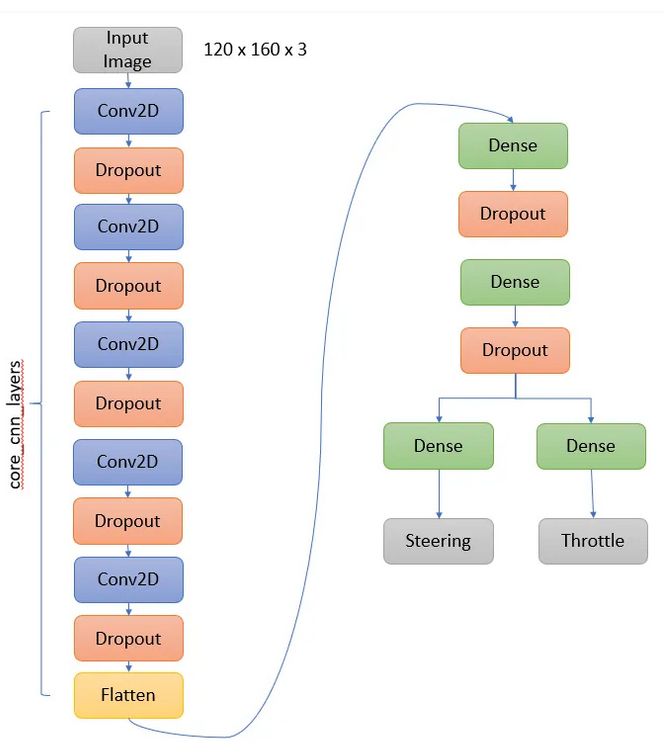}\label{fig:network}
    \caption{Network Structure} 
\end{figure}

\section{Results}

Our proposed scenario for evaluating our model can be seen in the Figure 3. It is approximately 13 meters long in a 5x5 meters area. After collecting the dataset and training our model, we set our vehicle to perform at it and on average we managed to make it complete the track in less than 20 seconds, giving us a pace of approximately 0.65 meters per second. 

\begin{figure}[httb]
    \centering
           \includegraphics[scale=0.1]{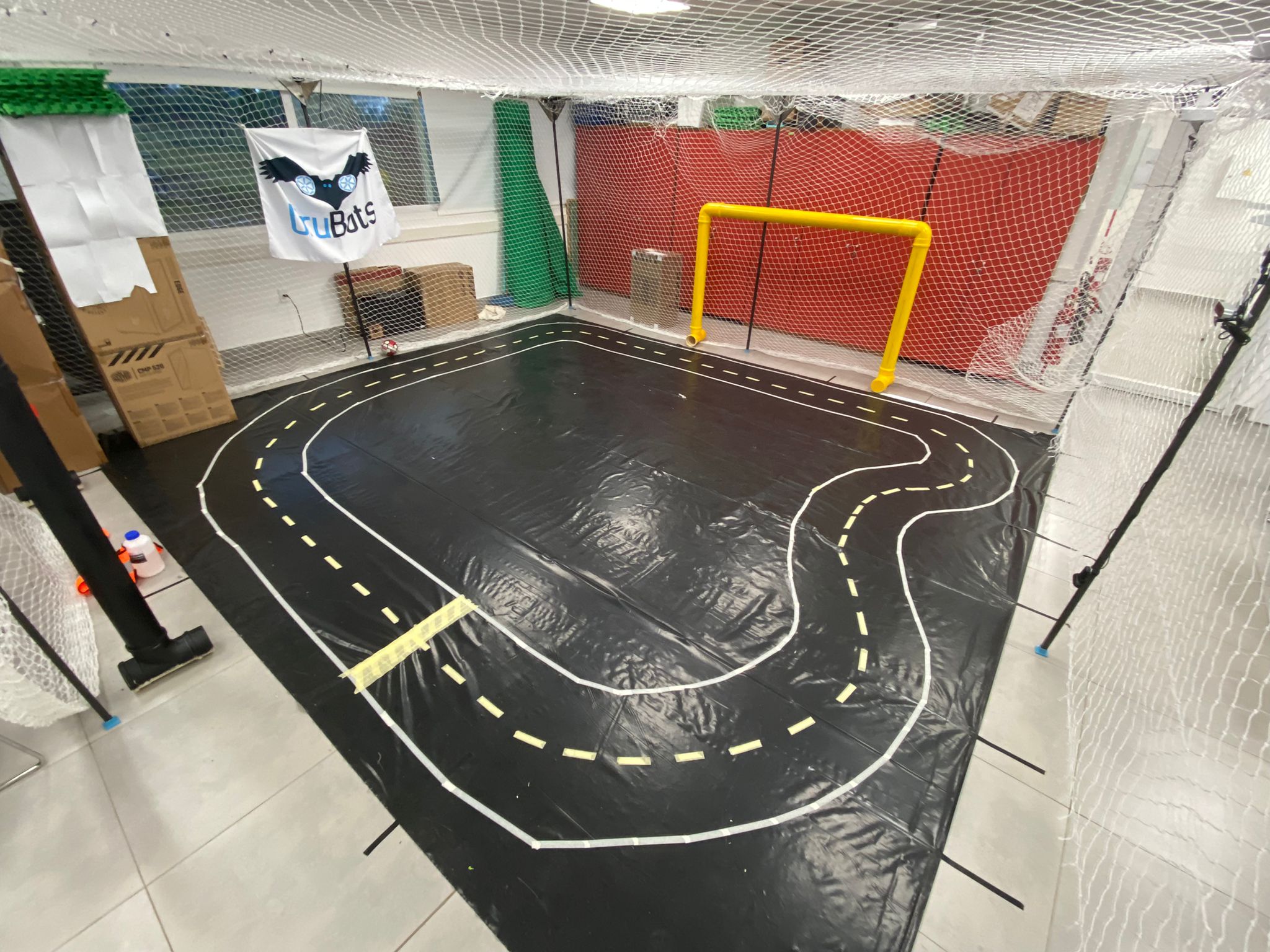}\label{fig:scenario}
    \caption{Track Scenario used to validate our vehicle} 
\end{figure}

\section{Conclusion}

Overall, we conclude that our proposed vehicle is capable to perform autonomous driving in race track and even avoid obstacles. We believe that our approach based on a learning model, advanced sensing and embedded processing met the requirements to perform at the 2024 FIRA RoboWorld Cup and help the development of this field of study and the competition itself.

\vspace{30pt} 
\begin{minipage}{\linewidth}

\end{minipage}



\begin{thebibliography}{8}
\bibitem{2}
Amirmohammad Zarif, Amirmahdi Zarif y Soroush Sadeghnejad (2024). Autonomous Cars Challenge Physical (Pro)
\end{thebibliography}

\end{document}